\documentclass[12pt,journal,compsoc]{IEEEtran}
 \usepackage{amsfonts} 
 \usepackage{amsmath}
 \usepackage{amsthm}
 \usepackage{pifont}
 \usepackage{authblk}
\usepackage{enumitem}
 \usepackage{graphicx}
 \usepackage{algorithm}
 \usepackage{algpseudocode}
\usepackage{textcomp}
\usepackage{gensymb}
\usepackage{color}
 
\RequirePackage{lineno}
\usepackage{subfig}
\usepackage{amssymb}
\usepackage{multirow}
\usepackage{subfig}
\usepackage{setspace}
\usepackage{url}
\usepackage{color}

\newtheorem{thm}{Theorem}
\newtheorem{lem}{Lemma}

\newtheorem*{defi*}{Definition}

\begin{document}

\title{Sparse Representation Classification Beyond $\ell 1$ Minimization and the Subspace Assumption}
\author{Cencheng Shen, Li Chen, Yuexiao Dong, Carey E. Priebe\thanks{Cencheng Shen is with Department of Applied Economics and Statistics at University of Delaware, Li Chen is with Intel, Yuexiao Dong is with Department of Statistical Science at Temple University, and Carey~E.~Priebe is with Department of Applied Mathematics and Statistics at Johns Hopkins University (email: shenc@udel.edu; lichen.jhu1@gmail.edu; ydong@temple.edu;  cep@jhu.edu). This work was partially supported by Johns Hopkins University Human Language Technology Center of Excellence, the XDATA program of the Defense Advanced Research Projects Agency administered through Air Force Research Laboratory contract FA8750-12-2-0303 and the SIMPLEX program through SPAWAR contract N66001-15-C-4041, and the National Science Foundation Division of Mathematical Sciences award DMS-1712947. This paper was presented in part at Joint Statistical Meeting and ICML Learning and Reasoning with Graphs workshop. The authors thank the editor and reviewer for their constructive and valuable comments that lead to significant improvements of the manuscript.}}

\IEEEtitleabstractindextext{%
\begin{abstract}
The sparse representation classifier (SRC) has been utilized in various classification problems, which makes use of $\ell 1$ minimization and works well for image recognition satisfying a subspace assumption. In this paper we propose a new implementation of SRC via screening, establish its equivalence to the original SRC under regularity conditions, and prove its classification consistency under a latent subspace model and contamination. The results are demonstrated via simulations and real data experiments, where the new algorithm achieves comparable numerical performance and significantly faster. 
\end{abstract}

}

\maketitle

\section{Introduction}
\label{sec:intro}

Sparse coding is widely recognized as a useful tool in machine learning, thanks to the theoretical advancement in regularized regression and $\ell 1$ minimization \cite{OsbornePresnellTurlach2000a, OsbornePresnellTurlach2000b, DonohoHuo2001, EfronHastie2004, CandesTao2005, Donoho2006, CandesTao2006, CandesRombergTao2006}, as well as numerous classification and clustering applications in computer vision and pattern recognition \cite{WrightYangGaneshMa2009, WrightMa2010, YinEtAl2012, YangZhouGanesh2013, ElhamifarVidal2013, ChenShenVogelsteinPriebe2016}.

In this paper, we concentrate on the sparse representation classification (SRC), which is proposed in \cite{WrightYangGaneshMa2009} and exhibits state-of-the-art performance for robust face recognition. It is easy to implement, work well for data satisfying the subspace assumption (e.g. face recognition, motion segmentation, and activity recognition), is robust against data contamination, and can be extended to block-wise algorithm and structured data sets \cite{EldarMishali2009,EldarKuppingerBolcskei2010,ElhamifarVidal2012}. Given a set of training data $\mathcal{X}=[x_{1},\ldots,x_{n}] \in \mathbb{R}^{m \times n}$ with the corresponding known class labels $\mathcal{Y}=[y_{1},\ldots,y_{n}]$, the task here is to classify a new testing observation $x$ of unknown label. SRC identifies a small subset $\widehat{\mathcal{X}} \in \mathbb{R}^{m \times s}$ from the training data to best represent the testing observation, calculates the least square regression coefficients, and computes the regression residual for classification. Comparing to nearest-neighbor and nearest-subspace classifiers, SRC exhibits better finite-sample performance on face recognition and is argued to be robust against image occlusion and contamination.

Other steps being standard, the most crucial and time-consuming part of SRC is to extract the appropriate sparse representation for the testing observation. Among all possible representations, the sparse representation $\widehat{\mathcal{X}}$ that minimizes the residual and the sparsity level $s$ often yields a better inference performance by the statistic principle of parsimony and bias-variance trade-off. By adding the $\ell 0$ constraint to the linear regression problem, one can minimize the residual and the sparsity level $s$ at the same time. As $\ell 0$ minimization is NP hard and unfeasible for large samples, $\ell 1$ minimization is the best substitute due to its computational advantage, which has a rich theoretical literature on exact sparsity recovery under various conditions \cite{DonohoHuo2001, DonohoElad2003, CandesTao2005, Donoho2006, CandesTao2006, CandesRombergTao2006}. Towards this direction, it is argued in \cite{WrightYangGaneshMa2009} that SRC is able to find the most appropriate representation and ensures successful face recognition under the subspace assumption: if data of the same class lie in the same subspace while data of different classes lie in different subspaces, then the sparse representation $\widehat{\mathcal{X}}$ identified by $\ell 1$ minimization shall only consist of observations from the correct class. Moreover, using $\ell 1$ minimization and assuming existence of perfect representation, \cite{ElhamifarVidal2013} derives a theoretical condition for perfect variable selection.

However, to achieve correct classification, the sparse representation $\widehat{\mathcal{X}}$ does not need to perfectly represent the testing observation, nor only selects training data of the correct class. Indeed, a perfect representation is generally not possible when the feature (or dimension) size $m$ exceeds the sample size $n$, while an approximate representation is often non-unique. A number of literature have also pointed out that neither $\ell 1$ minimization nor the subspace assumption are indispensable for SRC to perform well \cite{RigamontiBrownLepetit2011, ZhangYangFeng2011, ShiEriksson2011, ChiPorikli2013}. Intuitively, SRC can succeed whenever the sparse representation $\widehat{\mathcal{X}}$ contains some training data of the correct class, and the correct class can dominate the regression coefficients. It is not really required to recover the most sparse representation by $\ell 1$ minimization or achieve a perfect variable selection under the subspace assumption.

The above insights motivate us to propose a faster SRC algorithm and investigate its classification consistency. In Section~\ref{sec:background} we introduce basic notations and review the original SRC framework. Section~\ref{sec:main} is the main section: in Section~\ref{mainSec0} we propose a new SRC algorithm via screening and a slightly different classification rule; in Section~\ref{mainSec1} we compare and establish the equivalence between the two  classification rules under regularity conditions; then we prove the consistency of SRC under a latent subspace mixture model in Section~\ref{mainSec2}, which is further extended to contamination models and network models. Our results better explain the success and applicability of SRC, making it more appealing in terms of theoretical foundation, computational complexity and general applicability. The new SRC algorithm performs much faster than before and achieves comparable numerical performance, as supported by a wide variety of simulations and real data experiments on images and network graphs in Section~\ref{sec:numer}. All proofs are in Section~\ref{sec:pf}.

\section{Preliminary}
\label{sec:background}
\subsection*{Notations}
Let $\mathcal{X}=[x_{1},x_{2},\ldots,x_{n}] \in \mathbb{R}^{m \times n}$ be the training data matrix and $\mathcal{Y}=[y_{1},y_{2},\ldots,y_{n}] \in [K]^{n}$ be the known class label vector, where $m$ is the number of dimensions (or feature size), $n$ is the number of observations (or sample size), and $K$ is the number of classes with $[K]=[1, \ldots, K]$. Denote $(x,y) \in \mathbb{R}^{m} \times [K]$ as the testing pair and $y$ is the true but unobserved label.

As a common statistical assumption, we assume $(x,y),(x_1,y_1),\cdots,(x_n,y_n)$ are all independent realizations from a same distribution $F_{XY}$. A classifier $g_{n}(x,D_{n})$ is a function that estimates the unknown label $y \in [K]$ based on the training pairs $D_{n}=\{(x_1,y_1),\cdots,(x_n,y_n)\}$ and the testing observation $x$. For brevity, we always denote the classifier as $g_{n}(x)$, and the classifier is correct if and only if $g_{n}(x) = y$. Throughout the paper, 
we assume all observations are of unit norm ($\|x_{i}\|_{2}=1$) because SRC scales all observations to unit norm by default.

The sparse representation is a subset of the training data, which we denote as 
\begin{align*}
\widehat{\mathcal{X}}=[\widehat{x}_{1},\widehat{x}_{2},\ldots,\widehat{x}_{s}] \in \mathbb{R}^{m \times s},
\end{align*}
where each $\widehat{x}_{i}$ is selected from the training data $\mathcal{X}$, and $s$ is the number of observations in the representation, or the sparsity level.
Once $\widehat{\mathcal{X}}$ is determined, $\hat{\beta}$ denotes the $s \times 1$ least square regression coefficients between $\widehat{\mathcal{X}}$ and $x$, and the regression residual equals $\|x-\widehat{\mathcal{X}} \hat{\beta}\|_{2}$. For each class $k \in [K]$ and a given $\widehat{\mathcal{X}}$, we define
\begin{align*}
& \widehat{\mathcal{X}}_{k}=\{\widehat{x}_{i} \in \widehat{\mathcal{X}}, i=1,\ldots,s \ | \ y_{\widehat{x}_{i}}=k\} \\
& \widehat{\mathcal{X}}_{-k}=\{\widehat{x}_{i} \in \widehat{\mathcal{X}}, i=1,\ldots,s \ | \ y_{\widehat{x}_{i}} \neq k\}.
\end{align*}
Namely, $\widehat{\mathcal{X}}_{k}$
is the subset of $\widehat{\mathcal{X}}$ that contains all observations from class $k$, and $\widehat{\mathcal{X}}_{-k} = \widehat{\mathcal{X}} - \widehat{\mathcal{X}}_{k}$. Moreover, denote $\hat{\beta}_{k}$ as the regression coefficients of $\hat{\beta}$ corresponding to $\widehat{\mathcal{X}}_{k}$, and $\hat{\beta}_{-k}$ as the regression coefficients corresponding to $\widehat{\mathcal{X}}_{-k}$, i.e.,
\begin{align*}
\widehat{\mathcal{X}}_{k} \hat{\beta}_{k}+ \widehat{\mathcal{X}}_{-k} \hat{\beta}_{-k}= \widehat{\mathcal{X}} \hat{\beta}.
\end{align*}
Note that the original SRC in Algorithm~\ref{algSRC1} uses the class-wise regression residual $\|x-\widehat{\mathcal{X}}_{k} \hat{\beta}_{k}\|_{2}$ to classify. 

\subsection*{Sparse Representation Classification by $\ell 1$}
\label{ssec:src}
SRC consists of three steps: subset selection, least square regression, and classification via regression residual. Algorithm~\ref{algSRC1} describes the original algorithm: Equation~\ref{l1min} identifies the sparse representation $\widehat{\mathcal{X}}$ and computes the regression coefficients $\hat{\beta}$; then Equation~\ref{maa} assigns the class by minimizing the class-wise regression residual. In terms of computation time complexity, the $\ell 1$ minimization step requires at least $O(mns)$, while the classification step is much cheaper and takes $O(msK)$.

\begin{algorithm}
\caption{Sparse Representation Classification by $\ell 1$ Minimization and Magnitude Rule} 
\label{algSRC1}
\begin{algorithmic}[1]
\Statex \textbf{Input}: 
The training data matrix $\mathcal{X}$, the known label vector $\mathcal{Y}$, the testing observation $x$, and an error level $\epsilon$. 
\Statex

\Statex \textbf{$\ell 1$ Minimization:} For each testing observation $x$, find $\widehat{\mathcal{X}}$ and $\hat{\beta}$ that solves the $\ell 1$ minimization problem:
\begin{equation}
\label{l1min}
\hat{\beta}=\arg\min \|\beta\|_{1}    \text{ subject to } \|x-\widehat{\mathcal{X}}\beta\|_{2} \leq \epsilon.
\end{equation} 
\Statex

\Statex \textbf{Classification:} Assign the testing observation by minimizing the class-wise residual, i.e.,  
\begin{equation}
\label{maa}
g_{n}^{\ell 1}(x)=\arg\min_{k \in [K]} \|x - \widehat{\mathcal{X}}_{k} \hat{\beta}_{k}\|_{2}, 
\end{equation} 
break ties deterministically. We name this classification rule as the magnitude rule.
\Statex

\Statex \textbf{Output}: The estimated class label $g_{n}^{\ell 1}(x)$. 
\end{algorithmic} 
\end{algorithm}

The $\ell 1$ minimization step is the only computational expensive part of SRC. 
Computation-wise, there exists various greedy and iterative implementations of similar complexity, such as $\ell 1$ homotopy method \cite{OsbornePresnellTurlach2000a, OsbornePresnellTurlach2000b, EfronHastie2004}, orthogonal matching pursuit (OMP) \cite{Tropp2004, TroppGilbert2007}, augmented Lagrangian method \cite{YangZhouGanesh2013}, among many others. We use the homotopy algorithm for subsequent analysis and numerical comparison without delving into the algorithmic details, as most L1 minimization algorithms share similar performances as shown in \cite{YangZhouGanesh2013}.

Note that model selection is inherent to $\ell 1$ minimization or almost all variable selection methods, i.e., one need to either specify a tolerance noise level $\epsilon$ or a maximum sparsity level in order for the iterative algorithm to stop. 
The choice does not affect the theorems, but can impact the actual numerical performance and thus a separate topic for investigation \cite{Zhang2009,CaiWang2011}. In this paper we simply set the maximal sparsity level $s=\min\{n/\log(n),m\}$ for both the $\ell 1$ minimization here and the latter screening method in Section~\ref{mainSec0}, which achieves good empirical performance for both algorithms.

\section{Main Results}
\label{sec:main}

In this section, we present the new SRC algorithm, investigate its equivalence to the original SRC algorithm, prove the classification consistency under a latent subspace mixture model, followed by further generalizations. Note that one advantage of SRC is that it is applicable to both high-dimensional problems ($m \geq n$) and low-dimensional problems ($m < n$). Its finite-sample numerical success mostly lies in high-dimensional domains where traditional classifiers often fail. For example, the feature size $m$ is much larger than the sample size $n$ in all the image data we consider, and $m=n$ for the network adjacency matrices. The new SRC algorithm inherits the same advantage, and all our theoretical results hold regardless of $m$.

\subsection{SRC via Screening and Angle Rule}
\label{mainSec0}
The new SRC algorithm is presented in Algorithm~\ref{algSRC2}, which replaces $\ell 1$ minimization by screening, then assigns the class by minimizing the class-wise residual in angle. To distinguish with the \textbf{magnitude rule} of Algorithm~\ref{algSRC1}, we name Equation~\ref{al2} as the \textbf{angle rule}. Algorithm~\ref{algSRC2} has a better computation complexity due to the screening procedure, which simply chooses $s$ observations out of $\mathcal{X}$ that are most correlated with the testing observation $x$, only requiring $O(mn+n \log(n))$ in complexity instead of $O(mns)$ for $\ell 1$. 

The screening procedure has recently gained popularity as a fast alternative of regularized regression for high-dimensional data analysis. The speed advantage makes it a suitable candidate for efficient data extraction for extremely large $m$, and can be equivalent to $\ell 1$ and $\ell 0$ minimization under various regularity conditions \cite{FanLv2008,FanSamworthWu2009,WassermanRoeder2009,FanFengSong2011,GenoveseEtAl2012,KolarLiu2012}. In particular, setting the maximal sparsity level as $s=\max\{n/ \log(n),m\}$ is shown to work well for screening \cite{FanLv2008}, thus the default choice in this paper.

\begin{algorithm}
\caption{Sparse Representation Classification by Screening and Angle Rule} 
\label{algSRC2}
\begin{algorithmic}[1]
\Statex \textbf{Input}: 
The training data matrix $\mathcal{X}$, the known label vector $\mathcal{Y}$, and the testing observation $x$. 
\Statex

\Statex \textbf{Screening:} Calculate $\Omega=\{|x_{1}^{T} x|, |x_{2}^{T} x|, \cdots, |x_{n}^{T} x|\}$ ($^{T}$ is the transpose), and sort the elements by decreasing order. Take $\widehat{\mathcal{X}}=\{x_{(1)}, x_{(2)}, \ldots, x_{(s)}\}$ with $s=\min\{n/\log(n),m\}$, where $|x_{(i)}^{T}x|$ is the $i$th largest element in $\Omega$.
\Statex

\Statex \textbf{Regression:} Solve the ordinary least square problem between $\widehat{\mathcal{X}}$ and $x$. Namely, compute $\beta=\widehat{\mathcal{X}}^{-1} x$ where $\widehat{\mathcal{X}}^{-1}$ is the Moore-Penrose inverse.
\Statex

\Statex \textbf{Classification:} Assign the testing observation by 
\begin{equation}
\label{al2}
g_{n}^{scr}(x)=\arg\min_{k \in [K]} \theta(x, \widehat{\mathcal{X}}_{k} \hat{\beta}_{k}), 
\end{equation}
where $\theta$ denotes the angle between vectors. Break ties deterministically. We name this classification rule as the angle rule.
\Statex

\Statex \textbf{Output}: The estimated class label $g_{n}^{scr}(x)$. 
\end{algorithmic} 
\end{algorithm}  

\subsection{Equivalence Between Angle Rule and Magnitude Rule}
\label{mainSec1}
The angle rule in Algorithm~\ref{algSRC2} appears different from the magnitude rule in Algorithm~\ref{algSRC1}. For given sparse representation $\widehat{\mathcal{X}}$ and the regression vector $\hat{\beta}$, we analyze these two rules and establish their equivalence under certain conditions. 

\begin{thm}
\label{thm1}
Given $\widehat{\mathcal{X}}$ and $x$, we have $g_{n}^{\ell 1}(x)=g_{n}^{scr}(x)$ when either of the following conditions holds.
\begin{itemize}
\item $K=2$ and $\widehat{\mathcal{X}}$ is of full rank;
\item Data of different classes are orthogonal to each other, i.e., $\theta(\widehat{\mathcal{X}}_{y} \hat{\beta}_{y},\widehat{\mathcal{X}}_{k} \hat{\beta}_{k})=0$ for all $k \neq y$.
\end{itemize}
\end{thm}

These conditions are quite common in classification: binary classification problems are prevalent in many supervised learning tasks, and random vectors in high-dimensional space are orthogonal to each other with probability increasing to $1$ as the number of dimension increases \cite{Vershynin2017}.
Indeed, in all the multiclass high-dimensional simulations and experiments we run in Section~\ref{sec:sim1}, the two classification rules yield very similar classification errors. We chose to use the angle rule in the new SRC algorithm because it provides a direct path to classification consistency while the magnitude rule does not.

\subsection{Consistency under Latent Subspace Mixture Model}
\label{mainSec2}
To investigate the consistency of SRC, we first formalize the probabilistic setting of classification based on \cite{DevroyeGyorfiLugosiBook}. Let
\begin{align*}
(X,Y), (X_{1}, Y_{1}), \ldots, (X_{n}, Y_{n}) \stackrel{i.i.d.}{\sim} F_{XY}
\end{align*}
denote the random variables of the sample realizations $(x,y), (x_{1}, y_{1}), \ldots, (x_{n}, y_{n})$. The prior probability of each class $k$ is denoted by $\rho_{k} \in [0,1]$ with 
\begin{align*}
\sum_{k=1}^{K}\rho_{k}=1,
\end{align*}
and the probability error is defined by 
\begin{align*}
L(g_{n})=Prob(g_{n}(X) \neq Y).
\end{align*}
The classifier that minimizes the probability of error is called the Bayes classifier, whose error rate is optimal and denoted by $L^{*}$. The sequence of classifiers $g_{n}$ is consistent for a certain distribution $F_{XY}$ if and only if 
\begin{align*}
L(g_{n}) \rightarrow L^{*} \mbox{ as } n \rightarrow \infty.
\end{align*}

SRC cannot be universally consistent, i.e., there exists some distribution $F_{XY}$ such that SRC is not consistent. A simple example is a two-dimensional data space where all the data lie on the same line passing through the origin, then SRC cannot distinguish between them (as the normalized data are essentially a single point), whereas a simple linear discriminant without normalization is consistent. To that end, we propose the following model:
\begin{defi*}[Latent Subspace Mixture Model]
We say $(X,Y) \sim F_{XY} \in (\mathbb{R}^{m} \times [K])$ follows a latent subspace mixture model if and only if there exists a lower-dimensional continuously supported latent variable $U \in \mathbb{R}^{d}$ ($d \leq m$), and $m \times d$ matrices $W_{k}\in \mathbb{R}^{(m \times d)}$ for each $k \in [K]$ such that
\begin{align*}
X | Y =W_{Y}U.
\end{align*}
\end{defi*}
Namely, we observe a high-dimensional object $X$, and there exists a hidden low-dimensional latent variable $U$ and an unobserved class-dependent transformation $W_{Y}$. The latent subspace mixture model well reflects the original subspace assumption: data of the same class lie in the same subspace, while data of different classes lie in different subspaces. The subspace location is determined by $W_k$, and the model does not require a perfect linear recovery. Similar models have been used in a number of probabilistic high-dimensional analysis, e.g., probabilistic principal component analysis in \cite{BishopTipping1999}. Note that each subspace represented by $W_k$ does not need to be equal dimension. As long as $m$ denotes the maximum dimensions of all $W_k$, then the matrix representation can capture all lower-dimensional transformations. For example, let $K=2, m=3, d=2, U=(u_1,u_2)^T$, and \begin{eqnarray*}
    W_{1}=
    \begin{bmatrix}
    1       & 0  \\
    0       & 1  \\
    0       & 0  
\end{bmatrix} \qquad
    W_{2}=
    \begin{bmatrix}
    1       & -1  \\
    0       & 0  \\
    0       & 0 
\end{bmatrix}.
\end{eqnarray*} 
Then the subspace associated with class 1 is two-dimensional with $X|(Y=1) = (u_1,u_2,u_1+u_2)^T$, and the subspace associated with class 2 is one-dimensional with $X|(Y=2) = (u_1-u_2,0,0)^T$.

\begin{defi*}[The Angle Condition]
Under the latent subspace mixture model, denote 
\begin{align*}
\mathcal{W}=[W_{1} | W_{2} | \cdots | W_{K}] \in \mathbb{R}^{(m \times Kd)}
\end{align*}
as the concatenation of all possible $W_k$, and $\mathcal{W} / W_{k}$ denotes the same concatenation excluding $W_{k}$. We say $\mathcal{W}$ satisfies the angle condition if and only if
\begin{align}
\label{eq:span}
\mathrm{span}(W_{k}) \cap \mathrm{span}(\mathcal{W} / W_{k}) = \{0\}
\end{align}
for each $k \in [K]$.
\end{defi*}
Essentially, the condition states that the subspace of each class does not overlap with other subspaces from other classes, therefore testing data from one class cannot be perfectly represented by any linear combinations of the training data from other classes. The angle condition and the latent subspace mixture model allow data of the same class to be arbitrarily close in angle, while data of different classes to always differ in angle, which leads to the classification consistency of SRC.

\begin{thm}
\label{thm:main}
Under the latent subspace mixture model and $\mathcal{W}$ satisfying the angle condition, Algorithm~\ref{algSRC2} is consistent with $L^{*}$ being zero, i.e., 
\begin{align*}
L(g_{n}^{scr})  \stackrel{n\rightarrow \infty}{\rightarrow} L^{*} =0.
\end{align*}
\end{thm}

\subsection{Robustness against Contamination}
In feature contamination, certain features or dimensions of the data are contaminated or unobserved, thus treated as zero. Under the latent subspace mixture model, this can be equivalently characterized by imposing the contamination on the transformation matrix $W_k$, i.e., some entries of $W_k$ are $0$. By default, we assume there is no degenerate observation where all features are contaminated to $0$.

\begin{defi*}[Latent Subspace Mixture Model with Fixed Contamination]
Under the latent subspace mixture model, define $V_k$ as the $1 \times m$ contamination vector for each class $k$:
\begin{align*}
&V_{k}(j)=1 \mbox{ when $j$th dimension is not contaminated,}\\
&V_{k}(j)=0 \mbox{ when $j$th dimension is contaminated}.
\end{align*}
Then the contaminated random variable $X$ is
\begin{align*}
X|Y=diag(V_{Y})W_{Y}U,
\end{align*}
where $diag(V_{k})$ is an $m \times m$ diagonal matrix satisfying $diag(V_{k})(j,j)=V_{k}(j)$.
\end{defi*}

A more interesting contamination model is the following:
\begin{defi*}[Latent Subspace Mixture Model with Random Contamination]
Under the latent subspace mixture model, for each class $k$ define $V_k \in [0,1]^{1 \times m}$ as the contamination probability vector, and $Bernoulli (V_{Y})$ as the corresponding 0-1 contamination vector where $0$ represents the entry being contaminated. Then the contaminated random variable $X$ is
\begin{align*}
X|Y=diag(Bernoulli (V_{Y})) W_{Y}U.
\end{align*}
\end{defi*}
The two contamination models are very similar, except one being governed by a fixed vector while the other is being governed by a random process. 

\begin{thm}
\label{thm:robust}
Under the fixed contamination model, Algorithm~\ref{algSRC2} is consistent when 
\begin{align*}
\mathcal{W}_{V}=[diag(V_{1})W_{1}|\cdots|diag(V_{K})W_{K}]
\end{align*}
satisfies the angle condition.

Under the random contamination model, Algorithm~\ref{algSRC2} is consistent when 
\begin{align*}
\mathcal{W}_{V}=[diag(\emph{I}(V_{1}=1))W_{1}|\cdots|diag(\emph{I}(V_{K}=1))W_{K}]
\end{align*}
satisfies the angle condition, where $\emph{I}$ is the indicator function.
\end{thm}
Note that the notation $\emph{I}(V_{k}=1)$ represents a $0-1$ vector that applies the indicator function element-wise to class $k$ data, which has an entry of $1$ if and only if the respective dimension of class $k$ is un-contaminated. For example, let $K=3$, $m=6$, $d=1$, and \begin{align*}
I(V_1=1) = [1,1,1,0,0,0], \\
I(V_2=1) = [1,1,0,0,1,0], \\
I(V_3=1) = [1,1,0,0,0,1]. 
\end{align*}
Namely, for class $1$ data, the last three dimensions are randomly contaminated but not the first three dimensions; for class $2$ data, the first two and the fifth dimension are not contaminated; for class $3$ data, the first two and the last dimension are not contaminated. The theorem holds for the random contamination model when the common un-contaminated dimensions satisfy the angle condition, e.g., the first two dimensions in the above example.

Since $\mathcal{W}_{V}$ can be regarded as a projected version of $\mathcal{W}$ where the projection is enforced by the contamination, Theorem~\ref{thm:robust} essentially states that if the angle condition can still hold for a projected version of $\mathcal{W}$, then SRC is still consistent. If all dimensions can be randomly contaminated with non-zero probability, $\mathcal{W}_{V}$ becomes the empty matrix and the theorem no longer holds. This is because different subspaces may now overlap with each other simply chance, thus SRC cannot be as consistent as before. 

In practice, one generally has no prior knowledge about how exactly the data is contaminated. Thus Theorem~\ref{thm:robust} suggests that SRC can still perform well when the number of contaminated feature is relatively small or sparse among all features. The simulation in Section~\ref{sec:simu0} shows that under the two contamination models in Theorem~\ref{thm:robust}, SRC performs almost as well as the no-contamination case; whereas if all dimensions are allowed to be contaminated, SRC exhibits a much worse classification error.

\subsection{Consistency under Stochastic Block-Model}
SRC is shown as a robust vertex classifier in \cite{ChenShenVogelsteinPriebe2016}, exhibiting superior performance than other classifiers for both simulated and real networks. Here we prove SRC consistency for the stochastic block model~\cite{HollandEtAl1983,SussmanEtAl2012,Lei2015}, which is a popular network model commonly used for classification and clustering. Although the results are extend-able to undirected, weighted, and other similar graph models, for ease of presentation we concentrate on the directed and unweighted SBM.
\begin{defi*}[Directed and Unweighted Stochastic Block Model (SBM)]
Given the class membership $\mathcal{Y}$. A directed stochastic block model generates an $n \times n$ binary adjacency matrix $\mathcal{X}$ via a class connectivity matrix $V \in [0,1]^{K \times K}$ by Bernoulli distribution $\mathbf{B}(\cdot)$:
\begin{align*}
\mathcal{X}(i,j)=\mathbf{B}(V(y_i,y_j)).
\end{align*}
\end{defi*}

From the definition, the adjacency matrix produced by SBM is a high-dimensional object that is characterized by a low-dimensional class connectivity matrix. It is thus similar to the latent subspace mixture model.

\begin{thm}
\label{thm:sbm}
Denote $\rho \in [0,1]^{K}$ as the $1 \times K $ vector of prior probability, $\emph{I}_{\{Y=i\}}$ as the Bernoulli random variable of probability $\rho_i$, and define a set of new random variables $Q_k$ and their un-centered correlations $q_{kl}$ as
\begin{align*}
Q_{k}&= \sum_{i=1}^{K} \emph{I}_{\{Y=i\}} V(k,Y), \\
q_{kl}&= \frac{E(Q_k Q_l)}{\sqrt{E(Q_k^2)E(Q_l^2)}} \in [0,1]
\end{align*}
for all $k,l \in [K]$.

Then Algorithm~\ref{algSRC2} is consistent for SBM vertex classification when the regression coefficients are constrained to be non-negative and
\begin{align}
\label{eq:sbm}
q_{kl}^{2} \cdot \frac{E(Q_k^2)}{E(Q_l^2)} < \frac{E(Q_k)}{E(Q_l)}
\end{align}
for all $1 \leq k < l \leq K$.
\end{thm}
Equation~\ref{eq:sbm} essentially allows data of the same class to be more similar in angle than data of different classes, thus inherently the same as the angle condition for the latent subspace mixture model. Note that the theorem requires the regression coefficients to be non-negative. This is actually relaxed in the proof, which presents a more technical condition that maintains consistency while allowing some negative coefficients. Since network data are non-negative and SBM is a binary model, in practice the regression coefficients are mostly non-negative in either $\ell 1$ minimization or screening. Alternatively, it is also easy to enforce the non-negative constraint in the algorithm if needed \cite{Meinshausen2013}, which yields similar numerical performance.

\section{Numerical Experiments}
\label{sec:numer}
In this section we compare the new SRC algorithm by screening to the original SRC algorithm in various simulations and experiments. The evaluation criterion is the leave-one-out error: within each data set, one observation is hold out for testing and the remaining are used for training, do the classification, and repeat until each observation in the given data is hold-out once. The simulations show that SRC is consistent under both latent subspace mixture model and stochastic block model and is robust against contamination. The phenomenon is the same for real image and network data. Overall, we observe that Algorithm~\ref{algSRC2} performs very similar to Algorithm~\ref{algSRC1} in accuracy and achieves so with significantly better running time. Additional algorithm comparison is provided in Section~\ref{sec:sim1} to compare various different choices on the variable selection and classification steps. Note that we also compared two common benchmarks, the k-nearest-neighbor classifier and linear discriminant analysis for a number of network simulations and real graphs in \cite{ChenShenVogelsteinPriebe2016}, which shows SRC is significantly better than those common benchmarks and thus not repeated here.

\subsection{Image-Related Experiments}

\subsubsection{Latent Subspace Mixture Simulation}
\label{sec:simu0}
\begin{figure}
\centering
\subfloat[]{
\includegraphics[width=0.23\textwidth]{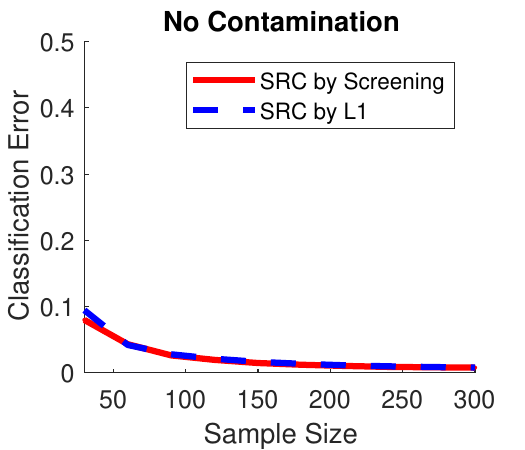}
}
\subfloat[]{
\includegraphics[width=0.23\textwidth]{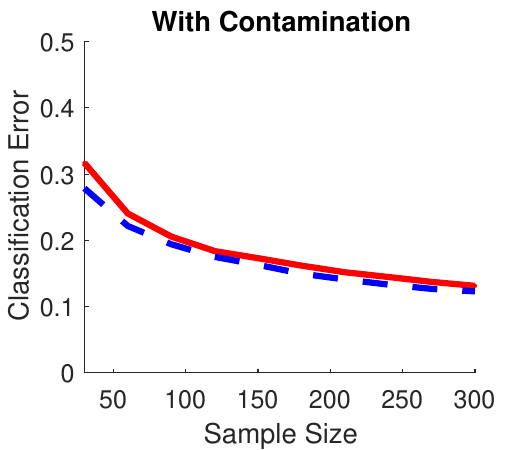}
}
\hfil
\subfloat[]{
\includegraphics[width=0.23\textwidth]{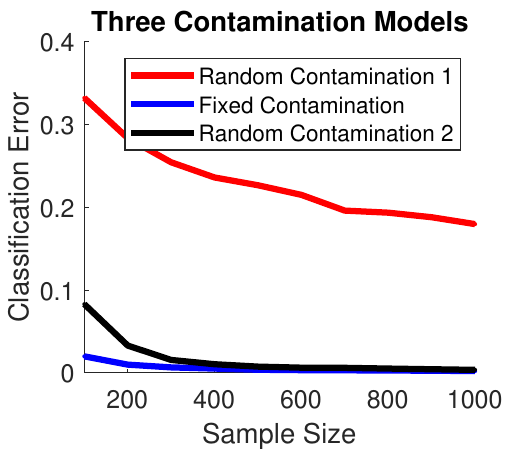}
}
\caption{SRC errors under Latent Subspace Mixture Model. The left and center panel compare algorithm~\ref{algSRC1} and algorithm~\ref{algSRC2} in no-contamination and $20\%$ random contamination data. The right panel compares SRC performance in three different contamination schemes, which shows SRC can perform very well when the contamination satisfies Theorem~\ref{thm:robust}. Note that the right panel only shows algorithm~\ref{algSRC2}, as the behavior is the same for algorithm~\ref{algSRC1}.}
\label{figSRC0}
\end{figure}

The model parameters are set as: $m=5$, $d=2$, $K=3$ with $\rho_{1}=\rho_{3}=0.3, \rho_{2}=0.4$. The $W_{k}$ matrices are:
\begin{eqnarray*}
    W_{1}=
    \begin{bmatrix}
    3       & 1  \\
    1       & 1  \\
    1       & 1 \\
    1       & 1\\
    \cdot   & \cdot 
\end{bmatrix} \qquad
    W_{2}=
    \begin{bmatrix}
    1       & 1  \\
    3       & 1  \\
    1       & 1 \\
    1       & 1\\
    \cdot   & \cdot 
\end{bmatrix} \qquad
    W_{3}=
    \begin{bmatrix}
    1       & 1  \\
    1       & 1  \\
    3       & 1 \\
    1       & 1\\
    \cdot   & \cdot 
\end{bmatrix},
\end{eqnarray*}
which satisfies the angle condition in Theorem~\ref{thm:main}.
We generate sample data $(\mathcal{X},\mathcal{Y})$ for $n=30,60,\ldots,300$, compute the leave-one-out error, then repeat for $100$ Monte-Carlo replicates and plot the average errors in Figure~\ref{figSRC0}. The left panel has no contamination, while the center panel has $20\%$ of the features contaminated to $0$ per each observation. In both panels, Algorithm~\ref{algSRC2} is almost as good as Algorithm~\ref{algSRC1}.

In the right panel of Figure~\ref{figSRC0}, we show how different contamination models may affect consistency. Consider three different contamination models: the same random contamination as in the center panel that sets each dimension to $0$ randomly for each observation; a fixed contamination that sets $20\%$ dimensions to $0$ for the same features throughout all data; and a second random contamination that never contaminate the first three dimensions (thus keeping the angle condition) with $20\%$ probability randomly sets each of the remaining dimensions to $0$ for each observation. The first contamination model is not guaranteed to be consistent, while the remaining two are guaranteed consistent by Theorem~\ref{thm:robust}. We use the same model setting as before except letting $m=10$ and growing sample size up-to $1000$ to better compare the convergence of the errors. Indeed, the results support the consistency theorem: for the fixed contamination or the random contamination satisfying the angle condition, SRC achieves almost perfect classification; while for the purely random contamination, the SRC error remains very high and is far from being optimal.

\subsubsection{Face and Object Images}
Next we experiment on two image data sets where original SRC excels at. The Extended Yale B database has $2414$ face images of $38$ individuals under various poses and lighting conditions \cite{GeorphiadesBelheumeurKriegman2001}, \cite{LeeHoKriegman2005}, which are re-sized to $32 \times 32$. Thus $m=1024$, $n=2414$, and $K=38$. The Columbia Object Image Library (Coil20) \cite{NeneNayarMurase1996a} consists of $400$ object images of $20$ objects under various angles, and each image is also of size $32 \times 32$. In this case $m=1024$, $n=400$, and $K=20$. 

The leave-one-out errors are reported in the first two rows of Table~\ref{t:table1}, and the running times are reported in the first two columns of Table~\ref{t:real2}. The new SRC algorithm is similar as the original SRC in error rate with a far better running time. Next we verify the robustness of Algorithm~\ref{algSRC2} against contamination. Figure~\ref{fig1} shows some examples of the image data, pre and post contamination. As the contamination rate increases from $0$ to $50\%$ of the pixels, the error rate increases significantly. Algorithm~\ref{algSRC2} still enjoys the same classification performance as the original SRC in Algorithm~\ref{algSRC1}, as shown in top two panels of Figure~\ref{figSRC2}.

\subsection{Network-Related Experiments}
\subsubsection{Stochastic Block Model Simulation}
\begin{figure}
\centering
\subfloat[]{
\includegraphics[width=0.23\textwidth]{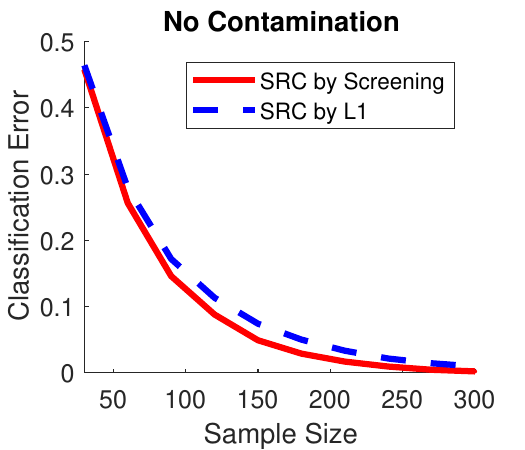}
}
\subfloat[]{
\includegraphics[width=0.23\textwidth]{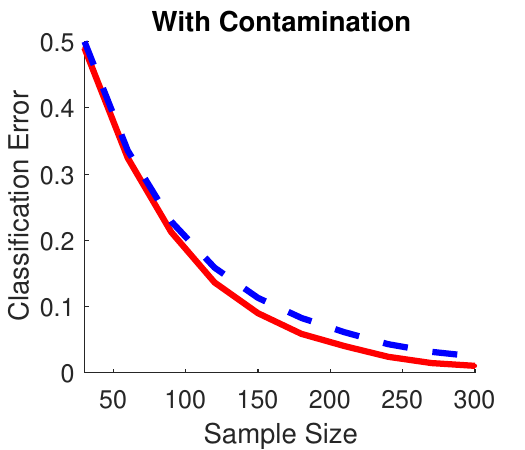}
}
\caption{SRC Errors under Stochastic Block Model. }
\label{figSRC1}
\end{figure}

\begin{figure}
\centering
\subfloat[]{
\includegraphics[width=0.4\textwidth]{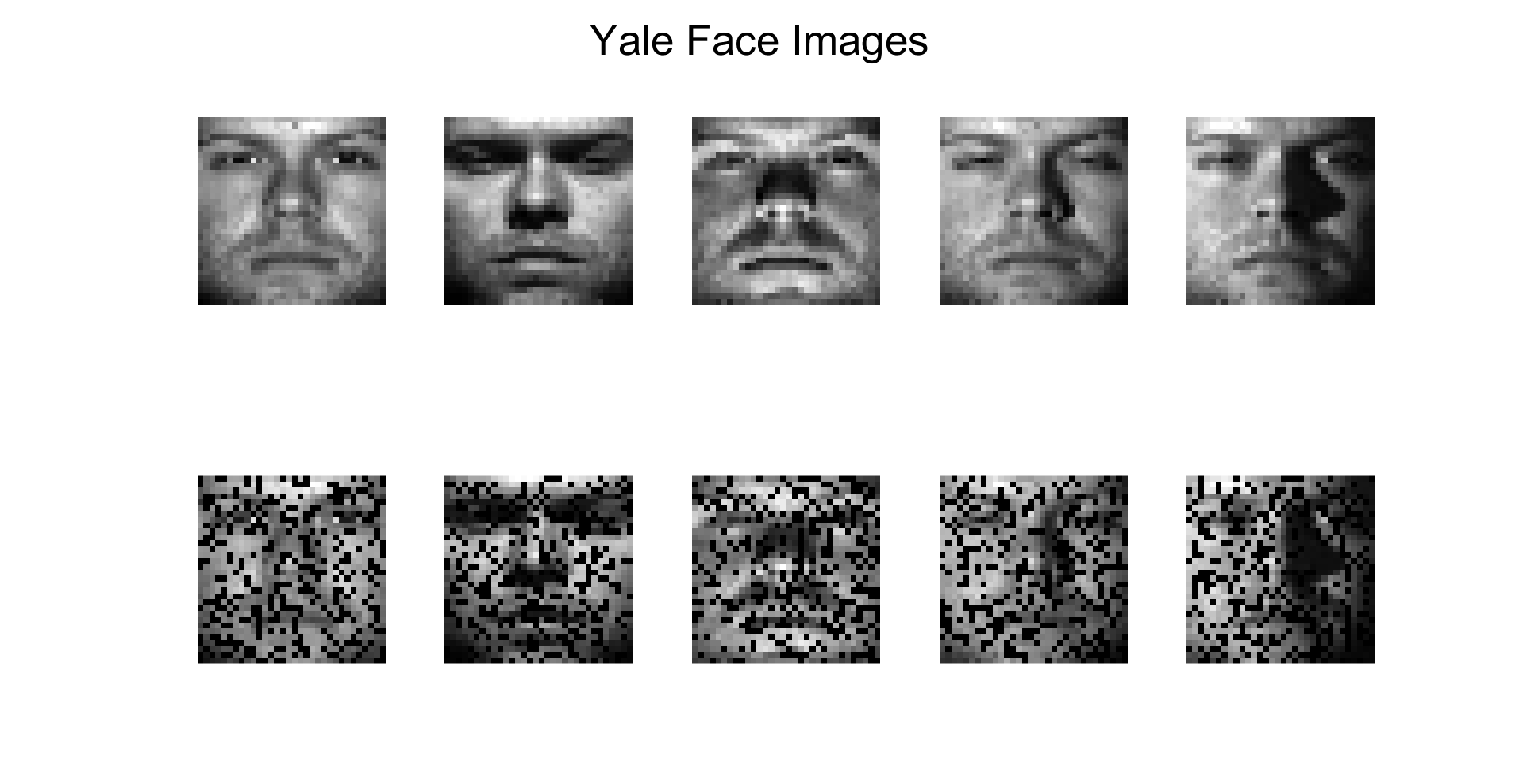}
}
\hfil
\subfloat[]{
\includegraphics[width=0.4\textwidth]{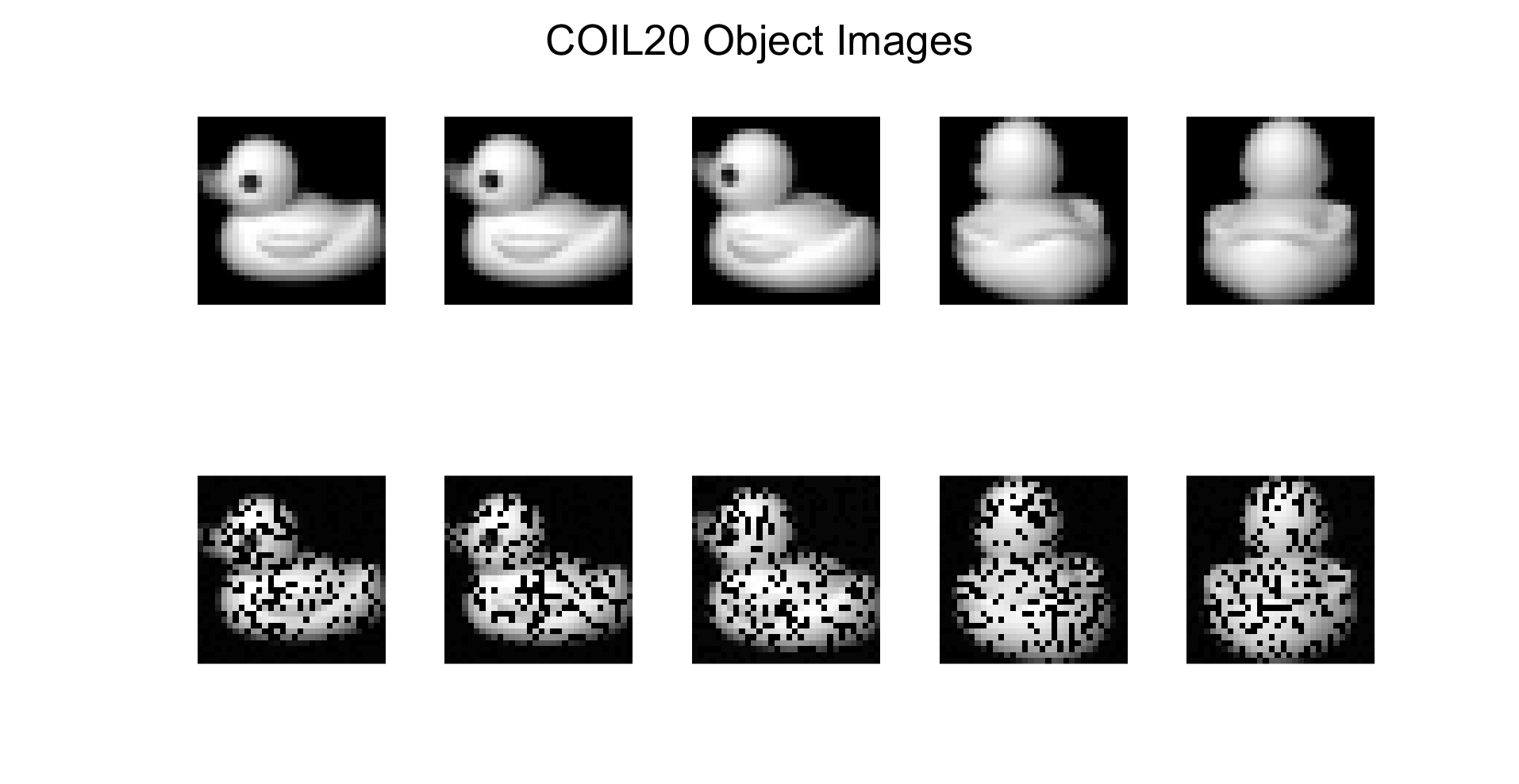}
}
\caption{Images with Contamination}
\label{fig1}
\end{figure}

Next we generate the adjacency matrix by the stochastic block model. We set $K=3$ with $\rho_{1}=\rho_{3}=0.3, \rho_{2}=0.4$, generate sample data $(\mathcal{X},\mathcal{Y})$ for $n=30,60,\ldots,300$, compute the leave-one-out error, then repeat for $100$ Monte-Carlo replicates and plot the average errors in Figure~\ref{figSRC1}. The class connectivity matrix $V$ is set to
\begin{eqnarray*}
    V=
    \begin{bmatrix}
    0.3       & 0.1 & 0.1 \\
    0.1       & 0.3 & 0.1  \\
    0.1       & 0.1 & 0.3
\end{bmatrix},
\end{eqnarray*}
which satisfies the condition in Theorem~\ref{thm:sbm}. The new SRC algorithm is similar to the original SRC algorithm in error rate; and both of them have very low errors, supporting the consistency result of Theorem~\ref{thm:sbm}.

\subsubsection{Article Hyperlinks and Neural Connectome}

In this section we apply SRC to vertex classification of network graphs. The first graph is collected from Wikipedia article hyperlinks \cite{PriebeMarchette2012}. A total of $1382$ English documents based on the 2-neighborhood of the English article ``algebraic geometry'' are collected, and the adjacency matrix is formed via the documents' hyperlinks. This is a directed, unweighted, and sparse graph without self-loop, where the graph density is $1.98\%$ (number of edges divided by the maximal number of possible edges). There are five classes based on article contents ($119$ articles in category class, $372$ articles about people, $270$ articles about locations, $191$ articles on date, and $430$ articles are real math). Thus, we have $m=n=1382$ and $K=5$. 

The second graph we consider is the electric neural connectome of Caenorhabditis elegans (\textit{C.elegans}) \cite{hall1991posterior,varshney2011structural,chen2015celegans}. The hermaphrodite \textit{C.elegans} somatic nervous system has over two hundred neurons, classified into $3$ classes: motor neurons, interneurons, and sensory neurons. The adjacency matrix is also undirected, unweighted, and sparse with density $1.32\%$. This is a relatively small data set where $m=n=253$ and $K=3$.

\begin{figure}
\subfloat[]{
\includegraphics[width=0.23\textwidth]{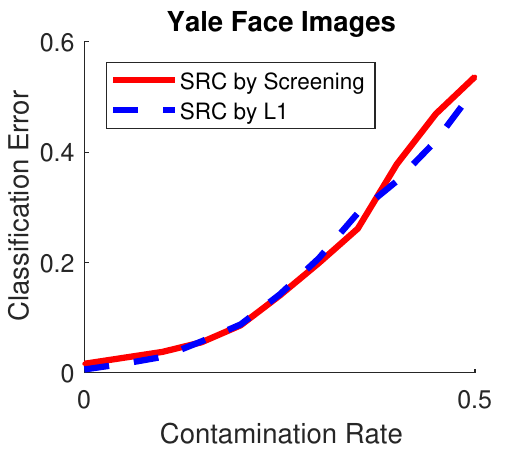}
}
\subfloat[]{
\includegraphics[width=0.23\textwidth]{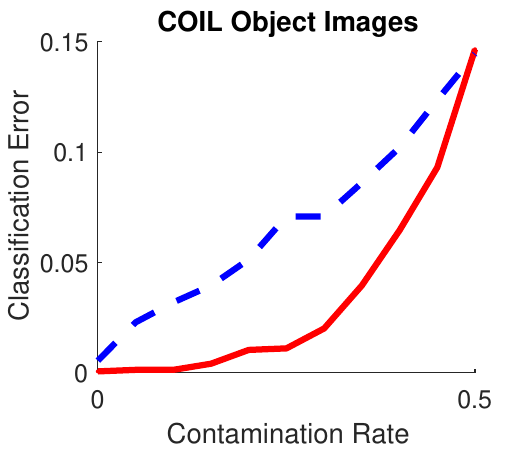}
}
\hfil
\subfloat[]{
\includegraphics[width=0.23\textwidth]{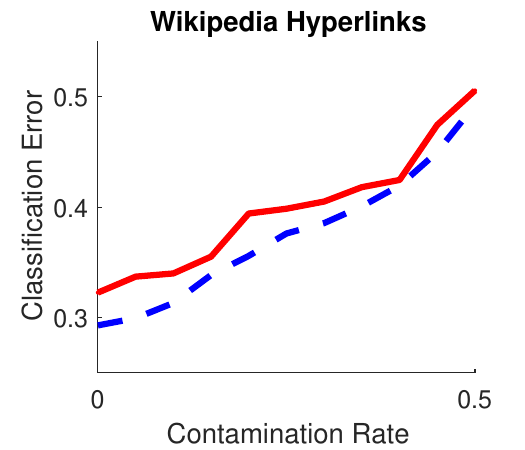}
}
\subfloat[]{
\includegraphics[width=0.23\textwidth]{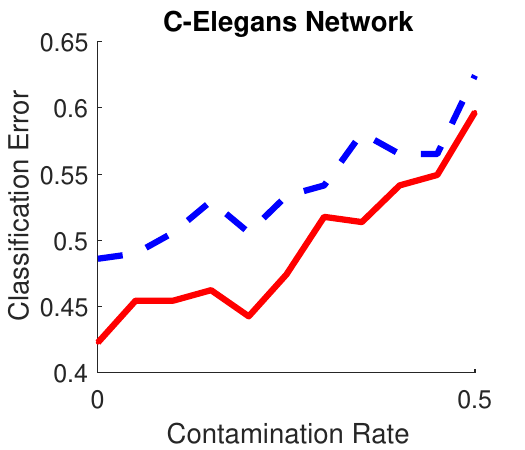}
}
\caption{SRC for Contaminated Real Data}
\label{figSRC2}
\end{figure}

The leave-one-out errors are reported in the first two rows of Table~\ref{t:table2}, the running times are reported in the last two columns of Table~\ref{t:real2}, and the contaminated classification performance are shown in the bottom panels of Figure~\ref{figSRC2}. The interpretation and performance curve are very similar to those of the image data, where Algorithm~\ref{algSRC2} is much faster without losing performance.

\begin{table*}
\centering
\caption{Running Time Comparison on Real Data (in seconds)}
\label{t:real2}%
\begin{tabular}{|c||c|c|c|c|c|c|}
\hline
Data & Yale Images & Coil Images & Wikipedia Graph & C-elegans Network \\
\hline
SRC by Screening & $72.7$ & $25.1$  & $32.3$ & $0.3$ \\
\hline
SRC by $\ell 1$  & $1101.5$ & $345.9$  & $573.7$ & $9.1$ \\
\hline
\end{tabular}
\end{table*}

\subsection{Experiments on Various Algorithmic Choices}
\label{sec:sim1}
The SRC algorithm can be implemented by other algorithm choices. For example, one may adopt a different variable selection technique for the first step in SRC, e.g., the OMP from \cite{TroppGilbert2007} is an ideal greedy algorithm to use, and screening followed by $\ell 1$ minimization may achieve better variable selection as argued in \cite{FanLv2008}. For the classifier rule, one naturally wonders what the practical difference between angle and magnitude rules is. To that end, we run the same simulation and experiments as above without contamination, via several different algorithmic choices. The results are summarized in Table~\ref{t:table1} for the image experiments and in Table~\ref{t:table2} for the network experiments. 

Overall, we do not observe much difference in accuracy regardless of these choices: 1. for the variable selection step, OMP is quite similar as $\ell 1$ homotopy, and screening $\circ \ \ell 1$ does not improve screening either, which implies that it suffices to simply use the fastest screening method for SRC. 2. for the classification rule, the angle and magnitude rules also behave either exactly the same or very similar, and one can be slightly better than another depending on the data in use. 

\begin{table*}
\centering
\caption{Leave-one-out error comparison for image-related simulations and real data. The first two rows correspond to the original and new SRC algorithms, and the remaining rows consider other algorithmic choices. For each data column, the best error rate is highlighted.}
\label{t:table1}%
\begin{tabular}{|c||c|c|c|}
\hline
Algorithm / Data & Latent Subspace & Yale Faces& Coil Objects  \\
\hline
$\ell 1$ with Magnitude (\textbf{Algorithm~\ref{algSRC1}}) & $1.33 \%$&  $\textbf{0.62} \%$  & $0.56 \%$\\
\hline
Screening with Angle (\textbf{Algorithm~\ref{algSRC2}}) & $\textbf{1.00} \%$ & $1.66 \%$ & $\textbf{0.01} \%$  \\
\hline
$\ell 1$ with Angle & $1.33 \%$ & $2.11 \%$ & $0.42 \%$  \\
\hline
Screening with Magnitude & $\textbf{1.00} \%$ & $0.79 \%$ & $\textbf{0.01} \%$   \\
\hline
OMP with Angle & $2.00 \%$ & $1.62 \%$ & $1.25 \%$  \\
\hline
OMP with Magnitude & $2.00 \%$ & $0.75 \%$ & $1.18 \%$   \\
\hline
Screening $\circ \ \ell 1$ with Angle & $1.33 \%$ & $3.07 \%$ & $0.21 \%$   \\
\hline
Screening $\circ \ \ell 1$ with Magnitude & $1.33 \%$ & $1.08 \%$ & $0.21 \%$  \\
\hline
\end{tabular}
\end{table*}

\begin{table*}
\centering
\caption{Leave-one-out error comparison for network-related simulations and real data. }
\label{t:table2}%
\begin{tabular}{|c||c|c|c|c|c|c|}
\hline
Algorithm / Data & SBM Simulation & Wikipedia& C-elegans\\
\hline
$\ell 1$ with Magnitude (\textbf{Algorithm~\ref{algSRC1}}) & $0.67 \%$  & $29.45 \%$ & $48.22 \%$ \\
\hline
Screening with Angle (\textbf{Algorithm~\ref{algSRC2}})& $\textbf{0.33} \%$  & $32.27 \%$ & $42.69 \%$ \\
\hline
$\ell 1$ with Angle & $0.67 \%$  & $\textbf{29.38} \%$ & $41.50 \%$ \\
\hline
Screening with Magnitude & $\textbf{0.33} \%$  & $32.27 \%$ & $45.06 \%$ \\
\hline
OMP with Angle & $2.33 \%$  & $30.97 \%$ & $40.32 \%$ \\
\hline
OMP with Magnitude & $2.33 \%$  & $32.13 \%$ & $46.25 \%$ \\
\hline
Screening $\circ \ \ell 1$ with Angle & $0.67 \%$  & $30.68 \%$ & $\textbf{39.53} \%$ \\
\hline
Screening $\circ \ \ell 1$ with Magnitude & $0.67 \%$  & $30.82 \%$ & $44.66 \%$ \\
\hline
\end{tabular}
\end{table*}

\section{Proofs}
\label{sec:pf}

\subsection*{Theorem~\ref{thm1} Proof}
To prove the theorem, we first state the following Lemma:
\begin{lem}
\label{lemma1}
Given the sparse representation $\widehat{\mathcal{X}}$ and the testing observation $x$, $g_{n}^{\ell 1}(x)=y$ if and only if $\| \widehat{\mathcal{X}}_{-y} \hat{\beta}_{-y}\|_{2} < \|\widehat{\mathcal{X}}_{-k} \hat{\beta}_{-k}\|_{2}$ for all classes $k \neq y$. And $g_{n}^{scr}(x)=y$ if and only if $\theta(x, \widehat{\mathcal{X}}_{y} \hat{\beta}_{y}) < \theta(x,\widehat{\mathcal{X}}_{k} \hat{\beta}_{k})$ for all classes $k \neq y$.
\end{lem}
\begin{proof}
This lemma can be proved as follows: the testing observation can be decomposed via $\widehat{\mathcal{X}}$ as
\begin{align*}
x &= \widehat{\mathcal{X}} \hat{\beta} +\epsilon \\
  &= \widehat{\mathcal{X}}_{k} \hat{\beta}_{k}+\widehat{\mathcal{X}}_{-k} \hat{\beta}_{-k}+\epsilon
\end{align*}
for any class $k$, where $\epsilon$ is the regression residual orthogonal to both $\widehat{\mathcal{X}}_{k} \hat{\beta}_{k}$ and $\widehat{\mathcal{X}}_{-k} \hat{\beta}_{-k}$. For the magnitude rule, $g_{n}^{\ell 1}(x)=y$ if and only if
\begin{align*}
&\|x-\widehat{\mathcal{X}}_{y} \hat{\beta}_{y}\| < \|x-\widehat{\mathcal{X}}_{k} \hat{\beta}_{k}\| \mbox{ for all } k \neq y \\ \Leftrightarrow &\|\widehat{\mathcal{X}}_{-y} \hat{\beta}_{-y}+\epsilon\| <\|\widehat{\mathcal{X}}_{-k} \hat{\beta}_{-k}+\epsilon\| \mbox{ for all } k \neq y \\
\Leftrightarrow &\|\widehat{\mathcal{X}}_{-y} \hat{\beta}_{-y}\| <\|\widehat{\mathcal{X}}_{-k} \hat{\beta}_{-k}\| \mbox{ for all } k \neq y, 
\end{align*}
where the last line follows because of the orthogonality of the regression residual $\epsilon$. For the angle rule, it is immediate that $g_{n}^{scr}(x)=y$ if and only if $\theta(x,\widehat{\mathcal{X}}_{y} \hat{\beta}_{y}) < \theta(x,\widehat{\mathcal{X}}_{k} \hat{\beta}_{k})$ for all $k \neq y$. This completes the proof of Lemma~\ref{lemma1}.
\end{proof}

Now we prove Theorem~\ref{thm1}: 
\begin{proof}
As $x =\widehat{\mathcal{X}}_{y} \hat{\beta}_{y}+\widehat{\mathcal{X}}_{-y} \hat{\beta}_{-y}+\epsilon$, it follows that
\begin{align}
\label{eq:main}
& \cos \theta(x, \widehat{\mathcal{X}}_{y} \hat{\beta}_{y}) = x^{T}\widehat{\mathcal{X}}_{y} \hat{\beta}_{y} / (\|x\|_{2} \|\widehat{\mathcal{X}}_{y} \hat{\beta}_{y}\|_{2}) \nonumber \\
  &=  (\|\widehat{\mathcal{X}}_{y} \hat{\beta}_{y}\|_{2}^{2}+ (\widehat{\mathcal{X}}_{-y} \hat{\beta}_{-y})^{T}\widehat{\mathcal{X}}_{y} \hat{\beta}_{y}) / \|\widehat{\mathcal{X}}_{y} \hat{\beta}_{y}\|_{2} \nonumber \\
	&=  \|\widehat{\mathcal{X}}_{y} \hat{\beta}_{y}\|_{2} + (\widehat{\mathcal{X}}_{-y} \hat{\beta}_{-y})^{T}\widehat{\mathcal{X}}_{y} \hat{\beta}_{y} / \|\widehat{\mathcal{X}}_{y} \hat{\beta}_{y}\|_{2} \nonumber \\
	&=  \|\widehat{\mathcal{X}}_{y} \hat{\beta}_{y}\|_{2} + \|\widehat{\mathcal{X}}_{-y} \hat{\beta}_{-y}\|_{2} \cdot \cos \theta(\widehat{\mathcal{X}}_{y} \hat{\beta}_{y}, \widehat{\mathcal{X}}_{-y} \hat{\beta}_{-y}). 
\end{align}
The first line expresses the angle via normalized inner products; the second line decomposes $x$, and $\epsilon$ is eliminated because it is orthogonal to both $\widehat{\mathcal{X}}_{y} \hat{\beta}_{y}$ and $\widehat{\mathcal{X}}_{-y} \hat{\beta}_{-y}$; the third line divides the square of $\|\widehat{\mathcal{X}}_{y} \hat{\beta}_{y}\|$ by itself; and the last line re-expresses the remainder term by angle. Using the above equation and Lemma~\ref{lemma1}, we show that the magnitude rule is the same as the angle rule under either of the following conditions:

\begin{itemize}
\item When $K=2$ and $\widehat{\mathcal{X}}$ is of full rank: first note that the angle between vectors satisfies $-1 \leq \cos \theta(\widehat{\mathcal{X}}_{y} \hat{\beta}_{y}, \widehat{\mathcal{X}}_{-y} \hat{\beta}_{-y}) \leq 1$. Moreover, when the representation is of full rank, $\widehat{\mathcal{X}}_{y} \hat{\beta}_{y}$ and $\widehat{\mathcal{X}}_{-y} \hat{\beta}_{-y}$ cannot be in the same direction, so the $\leq 1$ inequality becomes strict. Next, as there are only two classes, $\|\widehat{\mathcal{X}}_{-y} \hat{\beta}_{-y}\|$ becomes the representation of the other class, and $\cos \theta(\widehat{\mathcal{X}}_{y} \hat{\beta}_{y}, \widehat{\mathcal{X}}_{-y} \hat{\beta}_{-y})$ is the same for both classes. Assume $Y=1$ and $Y=2$ are the two classes. Then Equation~\ref{eq:main} simplifies to
\begin{align*}
&\cos \theta(x, \widehat{\mathcal{X}}_{1} \hat{\beta}_{1}) = a + b \cdot c \\
&\cos \theta(x, \widehat{\mathcal{X}}_{2} \hat{\beta}_{2}) = b + a \cdot c,
\end{align*}
where $a=\|\widehat{\mathcal{X}}_{1} \hat{\beta}_{1}\|, b=\|\widehat{\mathcal{X}}_{2} \hat{\beta}_{2}\|$, and $c=\cos \theta(\widehat{\mathcal{X}}_{y} \hat{\beta}_{y}, \widehat{\mathcal{X}}_{-y} \hat{\beta}_{-y})<1$. Without loss of generality, assume $g_{n}^{\ell 1}(x)=1$, which is equivalent to $a>b$ by Lemma~\ref{lemma1}. This inequality holds if and only if $\cos \theta(x, \widehat{\mathcal{X}}_{1} \hat{\beta}_{1}) > \cos \theta(x, \widehat{\mathcal{X}}_{2} \hat{\beta}_{2})$ or equivalently $\theta(x, \widehat{\mathcal{X}}_{1} \hat{\beta}_{1}) < \theta(x, \widehat{\mathcal{X}}_{2} \hat{\beta}_{2})$. Thus $g_{n}^{scr}(x)=1$ by Lemma~\ref{lemma1} on the angle rule. Therefore, the magnitude and the angle rule are the same.
\item When data of one class is always orthogonal to data of another class: under this condition, $\cos \theta(\widehat{\mathcal{X}}_{k} \hat{\beta}_{k}, \widehat{\mathcal{X}}_{-k} \hat{\beta}_{-k})=0$ for any $k$, so Equation~\ref{eq:main} simplifies to
\begin{align*}
\cos \theta(x, \widehat{\mathcal{X}}_{k} \hat{\beta}_{k})
& =\|\widehat{\mathcal{X}}_{k} \hat{\beta}_{k}\| \\
& = \|x-\widehat{\mathcal{X}}_{-k} \hat{\beta}_{-k}-\epsilon\|\\
&=\|x\|-\|\epsilon\|-\|\widehat{\mathcal{X}}_{-k} \hat{\beta}_{-k}\|,
\end{align*} 
where we used the fact that $\widehat{\mathcal{X}}_{k} \hat{\beta}_{k}$, $\widehat{\mathcal{X}}_{-k} \hat{\beta}_{-k}$ and $\epsilon$ are all pairwise orthogonal to each other in the third line. Since $\|x\|$ and $\|\epsilon\|$ are both fixed in the classification step, the class $y$ with the smallest $\|\widehat{\mathcal{X}}_{-y} \hat{\beta}_{-y}\|$ has the largest $\cos \theta(x, \widehat{\mathcal{X}}_{y} \hat{\beta}_{y})$ and thus the smallest angle. Therefore, $g_{n}^{\ell 1}(x)=g_{n}^{scr}(x)$ by Lemma~\ref{lemma1}.
\end{itemize}
\end{proof} 

\subsection*{Theorem~\ref{thm:main} Proof}
Recall that $(x,y)$ denotes the testing observation pair that is generated by $(X,Y)$. To prove the theorem, we state two more lemmas here: 
\begin{lem}
\label{lemma2}
Given a testing pair $(X,Y)$ generated under a latent subspace mixture model satisfying the angle condition. Denote $\widehat{\mathcal{X}}_{-Y}=[X_1,X_2,\ldots,X_s]$ as a collection of random variables $X_i$ with $Y_{i} \neq Y$, and $C=[c_1,c_2,\ldots,c_s]$ as nonzero a coefficient vector of size $s$. Then it holds that
\begin{align}
\label{eq:lemma2}
\min \{ \theta(X,\widehat{\mathcal{X}}_{-Y} \cdot C)\}>0.
\end{align}
\end{lem}
\begin{proof}
This lemma essentially states that the testing data $X$ cannot be perfectly explained by any linear combination of the training data from the incorrect class, i.e., $X \neq \widehat{\mathcal{X}}_{-Y} \cdot C$ for any $C$. Note that $C$ is not necessarily the regression coefficient, rather any arbitrary coefficients. Under the latent subspace mixture model, $X=W_{Y} U$ for the testing data and $X_{i}=W_{Y_{i}} U_{i}$ for the training data where each $Y_i \neq Y$. If there exists a vector $C$ such that $X=\widehat{\mathcal{X}}_{-Y}C$, then
\begin{align*}
& X=W_{Y} \cdot U = \widehat{\mathcal{X}}_{-Y}C = \sum_{i=1}^{s} W_{Y_{i}} \cdot U_{i} \cdot c_i \\
& \Leftrightarrow \mathrm{span}(W_{Y}) \cap \mathrm{span}(\mathcal{W}/W_{Y}) \neq \{0\},
\end{align*}
which contradicts the angle condition. 
\end{proof}

\begin{lem}
\label{lemma3}
Given a testing pair $(X,Y)$ generated under a latent subspace mixture model satisfying the angle condition. Denote $\{X_1, X_2, \ldots, X_{n}\}$ as a group of random variables that are independently and identically distributed as $X$ satisfying $Y_{i}=Y$ for all $i$. As $n \rightarrow \infty$ it holds that
\begin{align}
\label{eq:lemma3}
\theta(X, X_{(1)}) &\rightarrow 0.
\end{align}
where $X_{(1)}$ denotes the order statistic with the smallest angle difference to $X$.
\end{lem}

\begin{proof}
This lemma is guaranteed by the property of order statistics: for any $\epsilon>0$,
\begin{align*}
& Prob(\theta(X,X_i)< \epsilon)>0 \\
\Rightarrow & Prob(\theta(X,X_{(1)})< \epsilon) \stackrel{n \rightarrow \infty}{\rightarrow} 1.
\end{align*}
Therefore, as long as there are sufficiently many training data of the correct class, with probability converging to $1$ there exists one training observation of the same class that is sufficiently close to the testing observation.
\end{proof}

Now we prove Theorem~\ref{thm:main}:
\begin{proof}
To prove consistency, it suffices to prove that $g_{n}^{scr}(X)=Y$ asymptotically. By Lemma~\ref{lemma1} on the angle rule, it is equivalent to prove that for sufficiently large $n$, for every $k \neq y$ it holds that
\begin{align*}
\theta(X,\widehat{\mathcal{X}}_{Y} \hat{\beta}_{Y})  <\theta(X,\widehat{\mathcal{X}}_{k} \hat{\beta}_{k}).
\end{align*}

By Lemma~\ref{lemma2}, there exists a constant $\epsilon$ such that $\theta(X,\widehat{\mathcal{X}}_{k} \hat{\beta}_{k})> \epsilon$ regardless of $n$. By Lemma~\ref{lemma3}, as sample size increases, with probability converging to $1$ it holds that $\theta(X,X_{(1)})< \epsilon$ for which $Y_{(1)}=Y$. Moreover, $X_{(1)}$ is guaranteed to enter the sparse representation, as by definition it enters the sparse representation $\widehat{\mathcal{X}}$ the first during screening. Since $Y_{(1)}=Y$, $X_{(1)}$ is part of $\widehat{\mathcal{X}}_{Y}$, and it follows that with probability converging to $1$,
\begin{align*}
\theta(X,\widehat{\mathcal{X}}_{Y} \hat{\beta}_{Y}) \leq \theta(X,X_{(1)}) <\epsilon < \theta(X,\widehat{\mathcal{X}}_{k} \hat{\beta}_{k})
\end{align*}
for all $k \neq Y$. Thus Algorithm~\ref{algSRC2} is consistent under the latent subspace mixture model and the angle condition.
\end{proof}

\subsection*{Theorem~\ref{thm:robust} Proof}
\begin{proof}
In both contamination models, it suffices to prove that $\mathcal{W}_{V}$ satisfying the angle condition is equivalent to $\mathcal{W}$ satisfying the angle condition, then applying Theorem~\ref{thm:main} yields consistency in both models. 

In the fixed contamination case, the $\mathcal{W}$ matrix actually reduces to $\mathcal{W}_{V}$, so the conclusion follows directly. For the random contamination case, observe that each $W_i$ can be decomposed into two disjoint components
\begin{align*}
& W_i = diag(\emph{I}(V_{i}=1))W_{i}+diag(\emph{I}(V_{i}<1))W_{i} \\
& \mathrm{span}\{diag(\emph{I}(V_{i}=1))W_{i}\} \cap \mathrm{span}\{diag(\emph{I}(V_{i}<1))W_{i}\} \\
& = \{0\}.
\end{align*}
Therefore, as long as the uncontaminated part
\begin{align*}
\mathcal{W}_{V}=[diag(\emph{I}(V_{1}=1))W_{1}|\cdots|diag(\emph{I}(V_{K}=1))W_{K}]
\end{align*}
satisfies the angle condition, the random matrix $\mathcal{W}$ always satisfies the angle condition. For example, say the first three dimensions of $\mathcal{W}$ are not contaminated while the remaining dimensions can be contaminated with nonzero probability. Then the first three dimensions satisfying Equation~\ref{eq:span} guarantees $\mathcal{W}$ also satisfying Equation~\ref{eq:span}, regardless of how the remaining dimensions of $\mathcal{W}$ are contaminated.
\end{proof}

\subsection*{Theorem~\ref{thm:sbm} Proof}
\begin{proof}
By Lemma~\ref{lemma1} on the angle rule, it suffices to prove that for sufficiently large sample size, it holds that
\begin{align*}
\theta(x,\widehat{\mathcal{X}}_{y} \hat{\beta}_{y})  <\theta(x,\widehat{\mathcal{X}}_{k} \hat{\beta}_{k})
\end{align*}
for every $k \neq y$.
Without loss of generality, assume $x$ is the testing adjacency vector of size $1 \times n$ from class $1$, $x'$ is a training adjacency vector of size $1 \times n$ also from class $1$, and $\{x_1,x_2,\ldots,x_s\}$ is any group of adjacency vectors not from class $1$. Similar to the proof of Theorem~\ref{thm:main},
it further suffices to prove that at sufficiently large $n$, it holds that
\begin{align*}
\cos\theta(x,x')  > \cos \theta(x,\sum_{i=1}^{s} c_s x_s)
\end{align*}
for any non-negative and non-zero vector $C=[c_1,\ldots,c_s]$.

First for the within-class angle:
\begin{align*}
&\cos \theta (x,x^{'})\\
&= \frac{\sum_{j=1}^{n} \mathcal{B}(V(1,Y_{j})V(1,Y_{j}))}{\sqrt{\sum_{j=1}^{n} \mathcal{B}(V(1,Y_{j})) \sum_{j=1}^{n}\mathcal{B}(V(1,Y_{j}))}}\\
&= \frac{\sum_{j=1}^{n} \mathcal{B}(V(1,Y_{j})V(1,Y_{j}))/n}{\sqrt{\sum_{j=1}^{n} \mathcal{B}(V(1,Y_{j}))/n \sum_{j=1}^{n}\mathcal{B}(V(1,Y_{j}))/n}}\\
&\stackrel{n \rightarrow \infty}{\rightarrow} \frac{ \sum_{i=1}^{K}\rho_i V(1,i)V(1,i)}{\sum_{i=1}^{K}\rho_i V(1,i)}\\
&= \frac{ E(Q_1^2)}{E(Q_1)}.
\end{align*}

For the between-class angle:
\begin{align*}
&\cos \theta (x,\sum_{i=1}^{s}c_{i}x_{i})\\
&= \frac{\sum_{j=1}^{n}\sum_{i=1}^{s}c_{i} \mathcal{B}(V(1,Y_{j})V(y_{i},Y_{j}))}{\sqrt{\sum_{j=1}^{n} \mathcal{B}(V(1,Y_{j})) \sum_{j=1}^{n}\{\sum_{i=1}^{s}c_{i}\mathcal{B}(V(y_{i},Y_{j}))\}^2}}\\
&\stackrel{n \rightarrow \infty}{\rightarrow} \sum_{i=1}^{s}c_{i} E(Q_1 Q_{y_i}) / \\
& {\sqrt{E(Q_1)\{\sum_{i=1}^{s}c_{i}^2 E(Q_{y_i})+2\sum_{j>i}^{s}\sum_{i=1}^{s}c_i c_j E(Q_{y_i}Q_{y_j})\}}}\\
&\leq  \frac{\sum_{i=1}^{s}c_{i} E(Q_1 Q_{y_i})}{\sqrt{\sum_{i=1}^{s}c_{i}^2 E(Q_1) E(Q_{y_i})}}.
\end{align*}
The last inequality is because $\sum_{j>i}^{s}\sum_{i=1}^{s}c_i c_j E(Q_{y_i}Q_{y_j})>0$
for any non-negative coefficient vector $C$. Thus one can remove all these cross terms for the between-class angle and put a $\leq$ sign above. Since the network data is always non-negative and SBM is a binary model, the bulk of the regression vector is expected to be positive. In practice, the above inequality almost always holds for non-negative data even if a few regression coefficients turn out to be negative. 

Therefore, consistency holds when
\begin{align*}
&\cos \theta (x,x^{'}) > \cos \theta (x,\sum_{i=1}^{s}c_{i}x_{i}) \mbox{ \ \ \ a.s.} \\
\Leftarrow &\frac{ E(Q_1^2)}{E(Q_1)} \geq \frac{\sum_{i=1}^{s}c_{i} E(Q_1 Q_{y_i})}{\sqrt{\sum_{i=1}^{s}c_{i}^2 E(Q_1) E(Q_{y_i})}} \\
\Leftrightarrow & \frac{ E(Q_1^2)}{E(Q_1)} \geq \frac{\sum_{i=1}^{s}c_{i} q_{1 y_i}\sqrt{E(Q_1^2)E(Q_{y_i}^2)}}{\sqrt{\sum_{i=1}^{s}c_{i}^2 E(Q_1) E(Q_{y_i})}}\\
\Leftrightarrow & \sqrt{\sum_{i=1}^{s}c_{i}^2\frac{E(Q_{y_i})}{E(Q_1)}} >  \sum_{i=1}^{s}c_{i} q_{1 y_i} \sqrt{\frac{E(Q_{y_i}^2)}{E(Q_1^2)}}\\
\Leftarrow & \ q_{1k}^{2} \cdot \frac{E(Q_k^2)}{E(Q_1^2)} < \frac{E(Q_k)}{E(Q_1)} \mbox{ for all } k \neq 1,
\end{align*}
which is exactly Equation~\ref{eq:sbm} when generalized to arbitrary class $l$ instead of just class $1$. Note that this result is also used in Lemma 1 and Theorem 1 in \cite{ChenShenVogelsteinPriebe2016}. The condition can be readily verified on any given SBM model, and usually holds for models that are densely connected within-class and sparsely connected between-class.
\end{proof}

\bibliographystyle{ieeetr}
\bibliography{references}
\end{document}